\documentclass{article}

\PassOptionsToPackage{numbers, compress}{natbib}

\usepackage[final]{neurips_2020}
\usepackage[utf8]{inputenc} 
\usepackage[T1]{fontenc}    
\usepackage{hyperref}      
\usepackage{url}            
\usepackage{booktabs}       
\usepackage{amsfonts}     
\usepackage{nicefrac}       
\usepackage{microtype}     
\usepackage{graphicx}
\usepackage{enumitem}
\newcommand{{\AN}}{AttendNeXt}
\title{Faster Attention Is What You Need: A Fast Self-Attention Neural Network Backbone Architecture for the Edge via Double-Condensing Attention Condensers}

\author{Alexander Wong$^{1,2}$,  Mohammad Javad Shafiee$^{1,2}$, Saad Abbasi$^{1}$, Saeejith Nair$^{1}$, Mahmoud Famouri$^{2}$ \\
$^1$University of Waterloo, Waterloo, Ontario, Canada\\
$^2$DarwinAI,  Waterloo, Ontario, Canada 
}

\begin{document}

\maketitle

\begin{abstract}
With the growing adoption of deep learning for on-device TinyML applications, there has been an ever-increasing demand for efficient neural network backbones optimized for the edge. Recently, the introduction of attention condenser networks have resulted in low-footprint, highly-efficient, self-attention neural networks that strike a strong balance between accuracy and speed. In this study, we introduce a faster attention condenser design called double-condensing attention condensers that allow for highly condensed feature embeddings.  We further employ a machine-driven design exploration strategy that imposes design constraints based on best practices for greater efficiency and robustness to produce the macro-micro architecture constructs of the backbone.  The resulting backbone (which we name \textbf{AttendNeXt}) achieves significantly higher inference throughput on an embedded ARM processor when compared to several other state-of-the-art efficient backbones ($>10\times$ faster than FB-Net C at higher accuracy and speed and $>10\times$ faster than MobileOne-S1 at smaller size) while having a small model size ($>1.37\times$ smaller than MobileNetv3-L at higher accuracy and speed) and strong accuracy (1.1\% higher top-1 accuracy than MobileViT XS on ImageNet at higher speed).  These promising results demonstrate that exploring different efficient architecture designs and self-attention mechanisms can lead to interesting new building blocks for TinyML applications.
\end{abstract}

\section{Introduction}

Over the past decade, Deep Learning has made tremendous strides in achieving or even exceeding human-level performance in image perception tasks such as object segmentation, detection and classification. Many key emerging technologies such as autonomous driving, automated industrial inspection and augmented/virtual reality applications rely on the aforementioned perception tasks to work effectively. Most real world applications typically require on-device deployment of deep neural networks (DNNs) for the purposes of real-time inference, privacy and security. However, DNNs typically require significant on-board computational resources which makes them challenging to deploy on edge devices. Given the potential of Deep Learning on the edge, there has been significant effort in recent years on developing highly efficient DNNs for resource constrained devices~\cite{mehta2021mobilevit,wu2019fbnet,cai2019once,howard2019searching,mobileone}, with different strategies introduced to find a strong balance between efficiency and accuracy. For example, MobileNet~\cite{howard2019searching} and FB-Net~\cite{wu2019fbnet} architectures leveraged bottleneck inverted residuals blocks to achieve efficient networks. In another example, MobileOne~\cite{mobileone} architectures leveraged network re-parameterization to eliminate branches within the DNN architecture to achieve low latency inference. However, much of the aforementioned methods emphasized identifying efficient architectural patterns to discover optimal DNNs.  

Very recently, given the success in attention mechanisms in deep learning for improving a neural network's focus on relevant stimuli while attenuating irrelevant stimuli, there has been growing interest in exploring the utility of self-attention mechanisms for constructing efficient DNNs. One such effort yielded MobileViT~\cite{mehta2021mobilevit} architectures which leveraged self-attention within a vision transformer architectural design to achieve highly efficient DNNs. Another recent effort yielded the concept of attention condensers~\cite{wong2020tinyspeech,attendnet}, a novel self-attention mechanism that learns and produces a condensed embedding characterizing joint local cross-channel activation relationships. The use of attention condensers has led to small footprint, highly efficient self-attention DNNs called attention condenser networks that strike a strong balance between accuracy and latency~\cite{xu2022celldefectnet,attendnet, attendseg, wong2022tb}.  As such, the incorporation of self-attention for improving the balance of efficiency and accuracy in deep neural network architecture design holds tremendous promise but is only in its infancy of scientific exploration.

Motivated to further improve the efficiency of attention condenser networks, in this study we introduce a double-condensing attention condenser (DC-AC) self-attention mechanism, an enhancement of the attention condenser mechanism proposed in~\cite{wong2020tinyspeech}. The key enhancement to the original attention condenser design is the introduction of an additional condensation mechanism on the input features, which enables more condensed feature embedding to be learned for improved balance between efficiency and representational performance. In addition, we leverage a machine-driven design exploration strategy~\cite{wong2018ferminets} that imposes best practices design constraints for greater efficiency and robustness to produce macro-micro architecture constructs that leverage the proposed DC-AC mechanism, resulting in a fast backbone architecture we name \textbf{\AN}.  

The paper is organized as follows.  Section 2 details the machine-driven design exploration strategy used to create AttendNeXt, the architecture of the double-condensing attention condenser design, and the network architecture of AttendNeXt.  Section 3 presents the experimental setup of the study along with the experimental results  and discussion comparing AttendNeXt with state-of-the-art efficient backbones.

\section{Methods}

\subsection{Double-condensing attention condensers}
\begin{figure}
    \includegraphics[width=\linewidth]{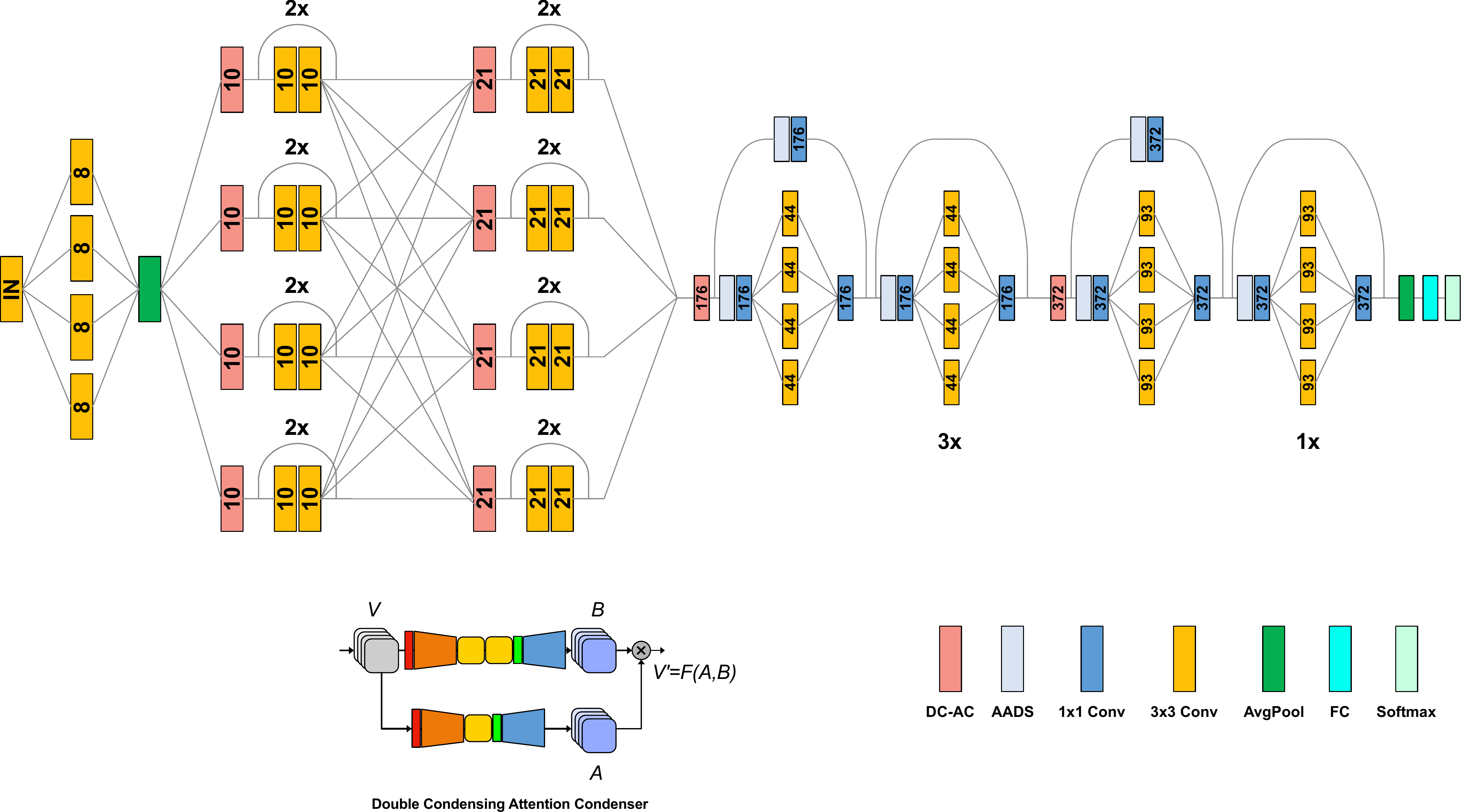}
    \caption{Proposed {\AN} architecture. DC-AC modules are comprised of condenser layers (orange), embedding layers (yellow) and expansion layers (blue).}
    \label{fig:attendnext}
\end{figure}

In this study, we introduce a double-condensing attention condenser (DC-AC) self-attention mechanism, an enhancement of the attention condenser mechanism first introduced in~\cite{wong2020tinyspeech} for incorporating efficient selective attention within a network architecture towards relevant stimuli. The architectural design of a DC-AC self-attention mechanism is shown in Figure~\ref{fig:attendnext}, and can be described as follows.  Fundamentally, DC-AC self-attention mechanisms are comprised of: 1) condenser layers, 2) embedding layers, and 3) expansion layers.  

More specifically, the condensation layers are responsible for condensing the input $V$ such that dimensionality is reduced in a manner that places greater emphasis on activations in close proximity to strong activations. This condensation allows for significantly reduced architectural and computational complexity while encouraging the characterization of strongly activated locations. The embedding layers are responsible for learning condensed embeddings characterizing joint local and cross-channel activation relationships.  This enables a rich understanding and characterization of what is important to focus on in the data while preserving lower architectural and computational complexity. The expansion layers are responsible for projecting the condensed embeddings to the appropriate dimensionality. The resulting self-attention values $A$ and feature embeddings $B$ are then leveraged to perform selective attention $V=F(A,B)$, which effectively leads to attenuation of irrelevant stimuli and direct attention to relevant stimuli. 

The key enhancement to the original attention condenser design is the introduction of an additional condensation mechanism on the input features $V$ (see top feature branch with output $B$), resulting in greater symmetry in condensation on both the feature branch and the attention branch in the DC-AC self-attention mechanism.  This symmetrical condensation within the self-attention mechanism in effect enables more condensed feature embeddings to be learned for improved balance between network efficiency and representational performance.  We further streamlined the attention condenser design by removing the scaling conducted during selective attention $F$, which was found to have limited effect in the new DC-AC design.  The proposed DC-AC mechanism is leveraged during machine-driven design exploration.

\subsection{Machine-driven design exploration}

In this study, generative synthesis~\cite{wong2018ferminets} is leveraged to perform machine-driven design exploration to determine the macro-micro architecture design of {\AN}. Generative synthesis is an generative architectural exploration process that discovers highly tailored DNN architectures based on operational requirements and constraints. The exploration process is formulated as a constrained optimization problem 

\begin{equation}
\mathcal{G}=\max_{\mathcal{G}}\mathcal{U}(\mathcal{G}(s)) \;\;\;\textrm{ subject to} \;\;\; 1_r(G(s))=1, \;\;\forall\in \mathcal{S}.
\end{equation}

where the goal is to find the optimal generator $G^\star(\cdot)$ which generates network architectures $N$ that maximizes a universal performance function $U$ (e.g.,~\cite{netscore}) under a given set of operational constraints formulated via an indicator function $\textbf{1}_r(\cdot)$. The optimization process is performed iteratively via the interplay between a generator $G^\star(\cdot)$, which is tasked with generating network architectures $N$ and an inquisitor $I$ responsible for assessing the performance of the $G$ via its generated architectures $N$.

We impose four key best practices design constraints via $\textbf{1}_r(\cdot)$ to identify macro- and micro- architecture designs for {\AN} that exhibit the desired balance between accuracy, architectural complexity, and computational complexity to produce a compact, low footprint, high performance neural network tailored for edge scenarios. These four constraints include:
\begin{enumerate}
    \item Encouraging columnar architecture designs with parallel columns to significantly reduce architectural and computational complexity.
    \item  Restricting the use of point-wise strided convolutions to mitigate the considerable information loss seen in many residual network designs~\cite{he2016deep,radosavovic2020designing}.
    \item Encouraging the use of anti-aliased downsampling (AADS)~\cite{zhang2019making} to improve network stability and robustness.
    \item Enforcing top-1 accuracy on ImageNet for a validation set to be above 75.8\% to ensure {\AN} performs at least as well as OFA-62, a compact architecture created via Once-For-All (OFA)~\cite{cai2019once}, a state-of-the-art neural architecture search method.
\end{enumerate}

\subsection{Network architecture}
\label{gen_inst}

Figure~\ref{fig:attendnext} demonstrates the {\AN} architecture designed via machine-driven exploration. A number of observations can be made from Figure~\ref{fig:attendnext}. First, it possesses a heterogeneous columnar design with different degrees of columnar interactions at different stages, thus striking a balance between accuracy and efficiency by learning representative yet disentangled embeddings. Secondly, by having more independent columns in the earlier stages and more columnar interactions at later stages, independent feature learning is amplified for lower levels of abstraction while more complex feature learning for higher levels of abstraction are catered for through the increased interactions amongst the columns. Thirdly, the widespread presence of AADS throughout the network architecture improves robustness and stability by making it more shift-invariant~\cite{zhang2019making}. Finally, we observe that DC-AC modules at different stages of the architectures increase efficiency of selective attention. Efficient selective attention results in improved representational efficiency through the use of condensed characterizations of joint spatial-channel activation relationships.

 \begin{figure}
     \centering
   \includegraphics[width=0.65\linewidth]{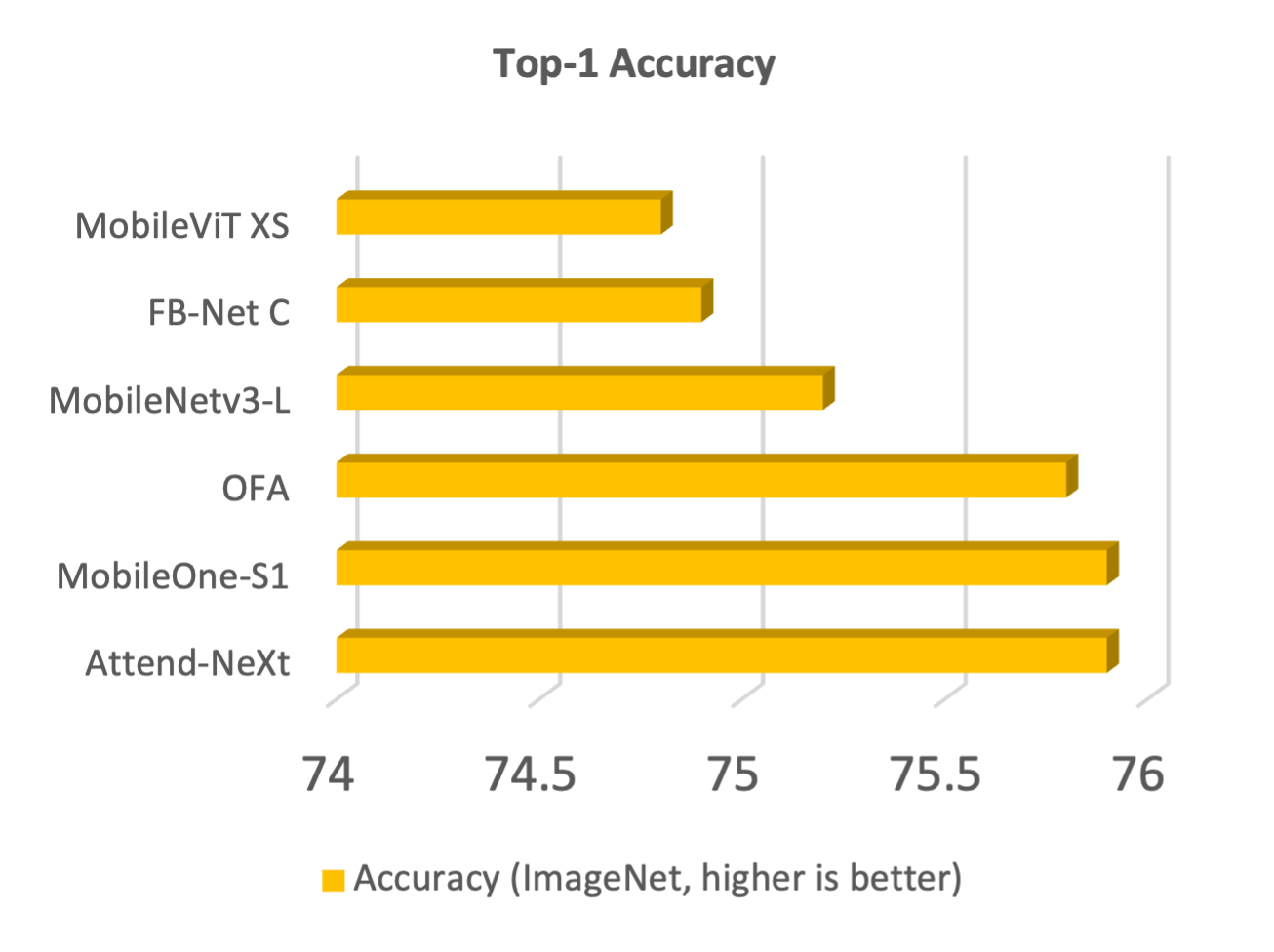}
     \caption{Top-1 accuracy (ImageNet) for {\AN} and state-of-the-art efficient networks. }
   \label{fig:acc}
 \end{figure}

\section{Results}
 
The efficacy of the proposed {\AN} architecture is evaluated on ImageNet dataset and is compared with the state-of-the-art efficient architectures (OFA-62 created via Once-For-All (OFA)~\cite{cai2019once}, FB-Net C~\cite{wu2019fbnet}, MobileViT XS~\cite{mehta2021mobilevit}, MobileOne-S1~\cite{mobileone}, and MobileNetv3-L~\cite{howard2019searching}) across three metrics: 1) Top-1 accuracy, 2) model size, and 3) relative inference throughput (with FB-Net C as baseline) on an embedded, low-power ARM Cortex A72 processor often used for edge computing applications.

 \begin{figure}

     \centering
 \hspace{-0.3in}\includegraphics[width=0.55\linewidth]{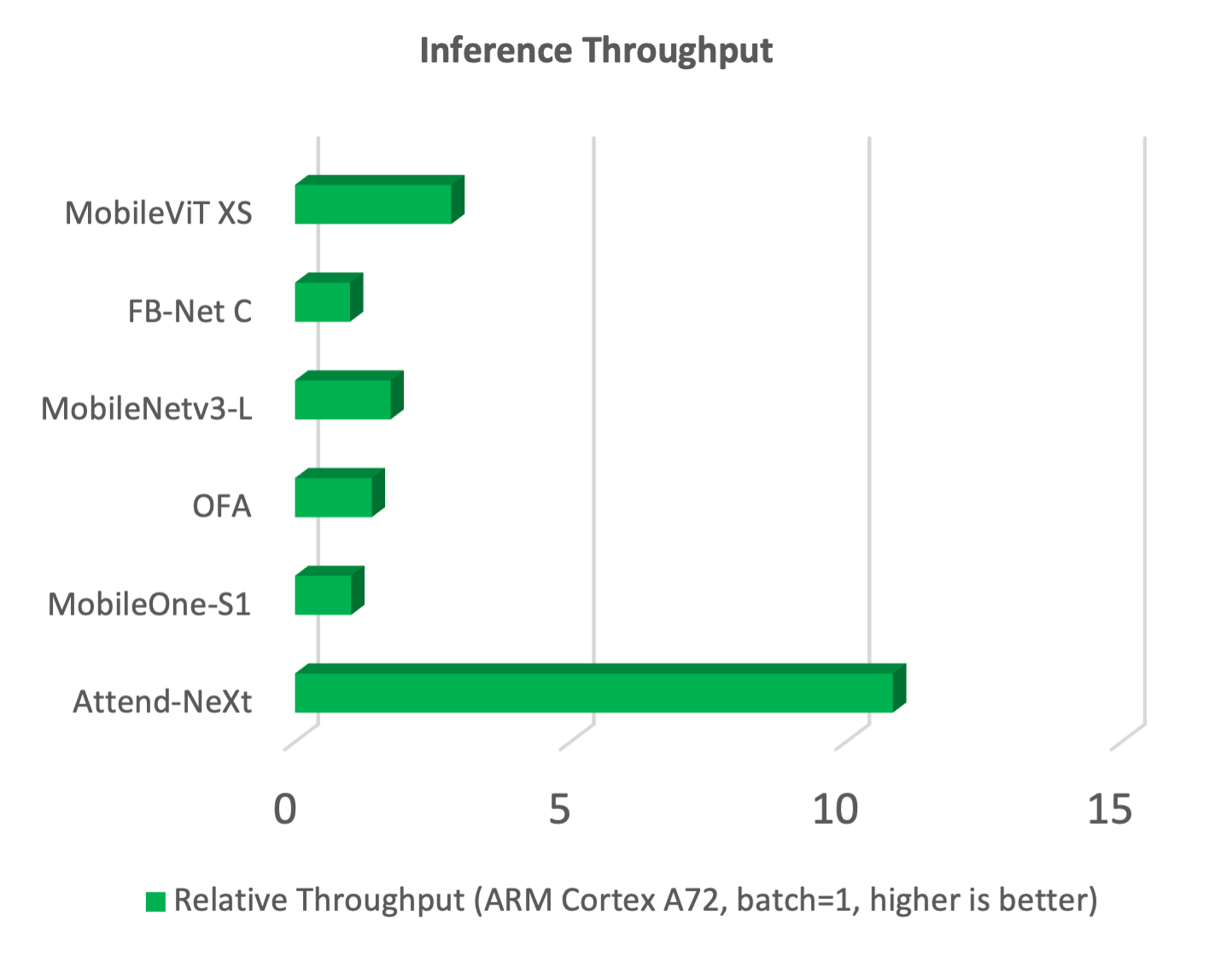}~\includegraphics[width=0.55\linewidth]{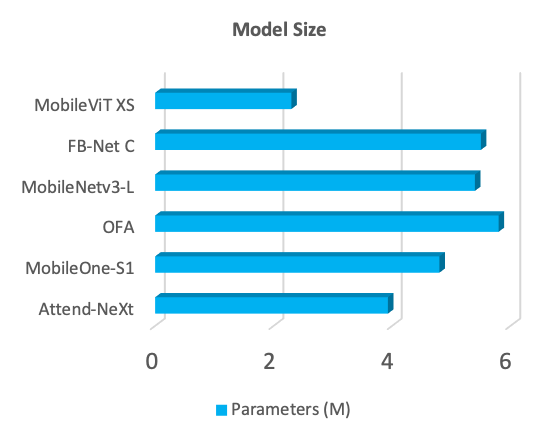}
     \caption{(a) Inference throughput on ARM Cortex A72. (b) Model size of \AN~and tested networks.}
     \label{fig:throughput}
 \end{figure}

Figure~\ref{fig:acc} shows the proposed {\AN} architecture achieved 75.8\% top-1 accuracy, outperforming MobileViT-XS and FB-Net C by $\sim$1\%. While {\AN} achieves slightly better top-1 accuracy than MobileNetV3-L and OFA-62, it significantly outperforms both architectures in inference throughput on ARM Cortex A72.  Figure~\ref{fig:throughput}(a) shows that {\AN} is $>10\times$ faster than FB-Net C and MobileOne-S1, $>6\times$ faster than OFA-62 and MobileNetv3-L, and $\sim4\times$ faster than MobileViT XS.  In terms of model size, Figure~\ref{fig:throughput}(b) shows that {\AN} is significantly smaller than OFA-62, MobileNetv3-L, FB-Net C, and MobileOne-S1 (e.g., {\AN} is $>1.47\times$ smaller than OFA-62).  While larger than MobileViT-XS, {\AN} is significantly faster on ARM Cortex A72 and significantly higher top-1 accuracy.  

These results demonstrate that {\AN} possesses a strong balance between accuracy, architectural complexity, and computational complexity, making such an architecture well-suited for TinyML applications on the edge.  Furthermore, it illustrates that efficient selective attention can enable a strong balance between network efficiency and representational power.  These promising results also demonstrate that exploring different efficient architecture designs and self-attention mechanisms can lead to interesting new building blocks for TinyML applications.

\section{Broader Impact}
The interest in TinyML (tiny machine learning) has grown significantly in recent years as one of the key driving factors towards widespread adoption of machine learning in industry and society.  TinyML is crucial for enabling real-time decision-making and predictive analytics on low-cost, low-power embedded devices, and holds tremendous potential that is currently being realized across industries ranging from manufacturing to energy to automotive to aerospace to healthcare.  The proliferation of TinyML can have considerable important socioeconomic implications that needs to be considered given its ability to enable machine learning across industries and applications.  Furthermore, the interest in attention mechanisms within deep learning has greatly increased in recent years, and its proliferation also has important implications given its increasing adoption within real-world applications.  The hope with explorations such as this study that introduces new building blocks such as double-condensing attention condensers is that it would provide new insights in advancing efforts in TinyML for greater adoption of machine learning towards ubiquity, as well as shed new insights into selective attention within a neural network context.

{\small
\bibliographystyle{unsrtnat}
\bibliography{main}
}

\end{document}